\documentclass[11pt]{article}
\usepackage[a4paper, margin=1in]{geometry}

\renewcommand{\baselinestretch}{1.31}
\newcommand{\linsp}{\renewcommand{\baselinestretch}{1.31}}
\newcommand{\linsps}{\renewcommand{\baselinestretch}{1.3}}

\usepackage[caption=false]{subfig}
\usepackage[usenames]{color}
\usepackage[figuresright]{rotating}
\usepackage{boxedminipage}
\usepackage[colorlinks,citecolor=blue,urlcolor=blue]{hyperref}
\usepackage{epstopdf}
\usepackage{natbib}
\usepackage{multirow}
\usepackage{enumerate}
\usepackage{enumitem}
\usepackage{verbatim}
\usepackage{bbm}
\usepackage{xcolor}
\usepackage{bm}
\usepackage{mathrsfs,color,dsfont}
\usepackage{algorithm,setspace}
\usepackage{algorithmic,xr}
\usepackage{threeparttable}

\renewcommand{\baselinestretch}{1.84}

\newcommand{\E}{\textsf{E}}

\newcommand{\mcN}{\mathcal{N}}
\newcommand{\mcT}{\mathcal{T}}
\newcommand{\mcF}{\mathcal{F}}
\newcommand{\mbR}{\mathbb{R}}

\newcommand{\bY}{\bm{Y}}
\newcommand{\bJ}{\bm{J}}
\newcommand{\bA}{\bm{A}}
\newcommand{\bW}{\bm{W}}
\newcommand{\bAz}{\bA_{\star}}
\newcommand{\bhA}{\widehat{\bA}}
\newcommand{\bepsilon}{\bm{\epsilon}}
\newcommand{\bM}{\bm{M}}
\newcommand{\bMz}{\bM_{\star}}
\newcommand{\bhM}{\widehat{\bM}}
\newcommand{\bhmu}{\widehat{\mu}}
\newcommand{\muz}{\mu_{\star}}

\newcommand{\bB}{\bm{B}}
\newcommand{\bC}{\bm{C}}

\newcommand{\bZ}{\bm{Z}}
\newcommand{\bG}{\bm{G}}
\newcommand{\bH}{\bm{H}}
\newcommand{\bhZ}{\widehat{\bZ}}
\newcommand{\bZz}{\bZ_{\star}}

\newcommand{\htheta}{\widehat{\theta}}
\newcommand{\bTheta}{\bm{\Theta}}
\newcommand{\bhTheta}{\widehat{\bTheta}}
\newcommand{\bThetaz}{\bTheta_{\star}}

\newcommand{\tGamma}{\widetilde{\Gamma}}

\newcommand{\bhinvThetaz}{\bTheta^{\ddagger}_{\star}}
\newcommand{\hatbhinvTheta}{\widehat{\bTheta}^{\ddagger}}
\newcommand{\binvTheta}{\bTheta^{\dagger}}
\newcommand{\binvThetaz}{\binvTheta_{\star}}
\newcommand{\hatbinvTheta}{\widehat{\bTheta}^{\dagger}}

\newcommand{\bU}{\bm{U}}

\newcommand{\bV}{\bm{V}}

\newcommand{\norm}[1]{\Vert#1\Vert}
\newcommand{\Norm}[1]{\left\Vert#1\right\Vert}

\newcommand{\Inner}[2]{\left\langle #1, #2 \right\rangle}
\newcommand{\abs}[1]{\vert#1\vert}
\newcommand{\Abs}[1]{\left\vert#1\right\vert}

\newcommand{\polylog}{\mathrm{polylog}}

\usepackage{amsthm,amsmath,amssymb,amsfonts, amsbsy, mathtools, epsfig}

\definecolor{DSgray}{cmyk}{0,1,0,0}

\def\AIC{\textsc{aic}}
\def\T{{ \mathrm{\scriptscriptstyle T} }}

\newtheorem{lemma}{Lemma}
\newtheorem{theorem}{Theorem}

\begin{document}
	\title{Matrix Completion under Low-Rank Missing Mechanism}
	\author{
		Xiaojun Mao
		\footnote{School of Data Science, Fudan University, Shanghai 200433, China. Email: maoxj@fudan.edu.cn} ~
		Raymond K. W. Wong
		\footnote{Department of Statistics, Texas A\&M University,
			College Station, Texas 77843, U.S.A. Email: raywong@stat.tamu.edu} ~
		Song Xi Chen
		\footnote{Department of Business Statistics and Econometrics, Guanghua School of Management and Center for Statistical Science, Peking University, Beijing 100651, China. Email: csx@gsm.pku.edu.cn} 
	}
	
	\date{}
	\maketitle
	\begin{abstract}
		
		Matrix completion is a modern missing data problem where both the missing structure and the underlying parameter are high dimensional. Although missing structure is a key component to any missing data problems, existing matrix completion methods often assume a simple uniform missing mechanism. In this work, we study matrix completion from corrupted data under a novel low-rank missing mechanism. The probability matrix of observation is estimated via a high dimensional low-rank matrix estimation procedure, and further used to complete the target matrix via inverse probabilities weighting. Due to both high dimensional and extreme (i.e., very small) nature of the true probability matrix, the effect of inverse probability weighting requires careful study. We derive optimal asymptotic convergence rates of the proposed estimators for both the observation probabilities and the target matrix.
		
	\end{abstract}
	\noindent
	\textbf{keywords:}
	Low-rank; Missing; Nuclear-norm; Regularization.
	
\section{Introduction}

The problem of recovering a high-dimensional matrix $\bAz\in\mathbb{R}^{n_1\times n_2}$ from very few
(noisy) observations of its entries is commonly known as matrix completion,
whose applications include, collaborative filtering, computer visions
and positioning.
From a statistical viewpoint, it is a high-dimensional missing data problem
where a high percentage of matrix entries are missing.
As in many missing data problems,
the underlying missing
mechanism plays an important role.
Most existing work \citep[e.g.,][]{Candes-Recht09,
	Keshavan-Montanari-Oh09, Recht11, Rohde-Tsybakov11,
	Koltchinskii-Lounici-Tsybakov11} adopt a uniform observation mechanism, where 
each entry has the same marginal probability of being observed. 
This leads to significant simplifications, and enables the domain
to move forward rapidly with various theoretical breakthroughs in the last decade.
However, the uniform mechanism 
is often unrealistic.
Recent works
\citep{Foygel-Shamir-Srebro11,Negahban-Wainwright12,
	Klopp14, Cai-Zhou16, Cai-Cai-Zhang16, Bi-Qu-Wang17, Mao-Chen-Wong19} 
have been devoted to relaxing such an restrictive assumption
by adopting other
missing structures.
The usage of these settings hinges on strong prior knowledge of the
underlying problems.
At a high level, many of them utilize some special forms of low-rank structures for missing mechanism.
For instance, \citet{Foygel-Shamir-Srebro11} and \citet{Negahban-Wainwright12}
both adopt a rank-1 structure based on the estimated marginal probabilities.
In this paper,
we aim at recovering the target matrix $\bAz$
under a flexible high-dimensional low-rank sampling structure.
This is achieved by a weighted empirical risk minimization,
with application of inverse probability weighting
\citep[e.g.,][]{Schnabel-Swaminathan-Singh16, Mao-Chen-Wong19} to
adjust for the effect of non-uniform missingness.

Data arising in many applications of matrix completion, such as
recommender systems, usually possesses complex 
``sampling" structure which is largely unknown. For a movie
recommender system, users tend to rate movies
that they prefer or dislike most, while often remain ``silent" to other movies. 
Another example of the complex sampling regime is in the online merchandising, where some
users may purchase certain items regularly without often rating them, but often evaluate
products that they rarely buy. 
Similar to the widely adopted model that ratings
are generated from a small number of hidden factors, it is reasonable to believe that
the missingness is also governed by a small and possibly different set of hidden factors,
which leads to a low-rank model the missing structure.

Inspired by generalized linear models, 
we model the probabilities of observation $\bThetaz=(\theta_{\star,ij})^{n_1,n_2}_{i,j=1}\in (0,1)^{n_1\times n_2}$  by a high-dimensional low-rank matrix
$\bMz=(m_{\star,ij})_{i,j=1}^{n_1,n_2}\in\mbR^{n_1\times n_2}$ through a known
function $f$.
That means, on the entry-wise level, we have $\theta_{\star,ij}=f(m_{\star,ij})$. 
In generalized linear models, 
the linear predictor $m_{\star,ij}$ is further modeled as a linear function of observed covariates.
However, to reflect difficulties to attain (appropriate and adequate) covariate information and the complexity in the modeling of $\bThetaz$ in some situations of the matrix completion, the predictor matrix $\bMz$ is assumed completely hidden in this study. Despite $\bMz$ being hidden, as demonstrated in this work, the low-rankness of $\bMz$ together with the high dimensionality of $\bMz$ allows both identification and consistent estimation of $\bThetaz$, which facilitates inverse probability weighting based matrix completion.
Motivated by the nature of matrix completion, we propose a novel parametrization $\bMz={\mu}_{\star}\bm{1}_{n_1}\bm{1}_{n_2}^{\T} + \bZz$
where $\bZz$ satisfies $\bm{1}_{n_1}^{\T} \bZz\bm{1}_{n_2}=0$. Our proposal extends the work of \citet{Davenport-Plan-Berg14}, which aims to solve a binary matrix completion problem and pursues a different goal. 
Compared with \citet{Davenport-Plan-Berg14}, the proposed method does not regularize
the estimation of $\muz$, but only regularize the nuclear norm of the estimation of $\bZz$, which require different algorithmic treatment to avoid bias caused by
the nuclear-norm penalty.  

There are three fundamental aspects
that set our work aside from the existing works of matrix completion as we consider:
(i) the high-dimensional probability matrix $\bm{\Theta}$, whose dimensions $n_1,n_2$ go to infinity in our asymptotic setting;
(ii) the diminishing lower bound of the observation probabilities (as $n_1,n_2$ go to infinity), and added issue to the instability of inverse probability weighting;
(iii) the effects of estimation error in inverse probability weighting to the matrix completion procedure.
Aspects (i) and (ii) are unique to our problem, and not found in the literature of missing data.
The work related to Aspect (iii) is sparse
in the literature of matrix completion. 
It is noted that \citet{Mao-Chen-Wong19} focused on a low-dimensional parametric modeling of inverse probability weighting with observable covariates.

We develop non-asymptotic upper bounds of the mean squared errors for the proposed estimators of the observation probabilities and the target matrix.
To sustain the convergence rate of the target matrix under the high-dimensionality of $\bMz$ and low levels of observation probabilities,
we propose to re-estimate $\bZz$ by
constraining the magnitude of its entries to a smaller threshold. 
Our analysis shows that
the proposed constrained inverse probability weighting estimator 
achieves the optimal rate (up to a logarithmic factor in estimation of target matrix). 
We also compare the inverse probability weighting based completion based on the proposed constrained estimation, with the completion based on direct weight trimming (or winsorization), a known practice in the conventional missing value literature \citep[e.g.,][]{Rubin01,Kang-Schafer07,Schafer-Kang08} and show that the constrained estimation has both theoretical and empirical advantages.

\section{Model and Method}\label{sec:model}

\subsection{General Setup}
Let $\bAz=(a_{\star,ij})_{i,j=1}^{n_1,n_2}\in \mbR^{n_1\times n_2}$ be an unknown high-dimensional matrix of interest, and $\bY=(y_{ij})_{i,j=1}^{n_1,n_2}$ be a contaminated version of $\bAz$ according to the following additive noise model:
\begin{equation}\label{eqn:datmod}
y_{ij} = a_{\star,ij} +\epsilon_{ij}, \quad \text{for $i=1,\dots, n_1;j=1,\dots,n_2$,} 
\end{equation}
where $\{\epsilon_{ij}\}$ are independently distributed random errors with zero mean and finite variance.
In the setting of matrix completion, only
a portion of $\{y_{ij}\}$ is observed.
For the $(i,j)$th entry, define the sampling indicator $w_{ij}=1$ if $y_{ij}$ is observed, and $0$ otherwise, and assume $\{\epsilon_{ij}\}$ are independent of $\{w_{ij}\}$.

As for the sampling mechanism, we adopt a Bernoulli model
where $\{w_{ij}\}$ are independent Bernoulli random variables
with observation probabilities $\{\theta_{\star,ij}\}$, collectively
denoted by a matrix
$\bThetaz=(\theta_{\star,ij})_{i,j=1}^{n_1,n_2}\in(0,1)^{n_1\times n_2}$.
Similar to generalized linear models,
the observation probabilities can be expressed in terms of an unknown matrix
$\bMz=(m_{\star, ij})_{i,j=1}^{n_1,n_2}\in \mbR^{n_1\times n_2}$
and a pre-specified monotone and
differentiable function $f:\mbR\to[0,1]$, i.e.,
$\theta_{\star,ij}=f(m_{\star,ij})$ for all $i,j$.
The matrix $\bMz$ plays the same role as a linear predictor in the generalized linear model,
while the function $f$ is an inverse link function.
Two popular choices of $f$
are inverse logit function
$g(m) = {e^m}/{(1 + e^m)}$ (logistic model)
and
the standard normal cumulative distribution function (probit model).

\subsection{Low-rank Modeling of \texorpdfstring{$\bAz$}{A0} and \texorpdfstring{$\bMz$}{M0}}

The above setup is general. Without additional assumption, it is virtually impossible to recover the hidden feature matrix $\bMz$ and also the target matrix $\bAz$.
A common and powerful assumption is that $\bAz$ is a low-rank matrix,
i.e.,~$\mathrm{rank}(\bAz) \ll \min\{n_1,n_2\}$.
Take the Yahoo!~Webscope data set (to be analyzed in Section \ref{sec:real}) as an example.
This data set contains a partially observed matrix of ratings from 15,400 users to 1000 songs,
and the goal is to complete the rating matrix.
The low-rank assumption reflects the belief that users' ratings are
generated by a small number of factors,
representing several standard preference profiles for songs.
This viewpoint has been proven useful in the modeling
of recommender systems \citep[e.g.,][]{Candes-Plan10,Cai-Candes-Shen10}. 

The
same idea could be adapted to the missing pattern,
despite that the factors that induce the missingness may be different from those
that generate the ratings.
To this end, we assume
$\bMz$
is also low-rank.
Next,
we consider decomposing
$\bm{M}_\star$ as 
\begin{equation} 
\bMz=\muz\bJ+\bZz \quad \hbox{ where} \quad 
\mbox{$\bm{1}_{n_1}^{\T}\bZz\bm{1}_{n_2}=0$}\label{eq:M-model}
\end{equation} 
with $\bm{1}_n$ being an $n$-vector of ones,
and $\bJ= \bm{1}_{n_1} \bm{1}_{n_2}^{\T}$.
Here $\muz$ is the mean of $\bMz$, i.e.,
$\muz=\bm{1}_{n_1}^{\T}\bMz\bm{1}_{n_2}/(n_1n_2)$.
Note that this decomposition
holds for any matrix $\bM$ by setting $\mu=(n_1n_2)^{-1}\bm{1}_{n_1}^{\rm T}\bm{M}\bm{1}_{n_2}$ and $\bm{Z}=\bm{M}-\mu\bm{J}$. Moreover, the decomposition is unique due to the constraint that $\bm{1}_{n_1}^{\T}\bZz\bm{1}_{n_2}=0$.
The key here is to reparametrize $\bMz$ in terms of $\mu_\star$ and $\bm{Z}_\star$,
which require different treatments in their estimation.
See Section \ref{sec:estTheta} for details.
Further, the low-rankness of $\bMz$ can be translated to the low-rankness of $\bZz$.

We note that the rank of $\bMz$ is not the same as
that of $\bThetaz$ due to the nonlinear transformation $f$.
In general, the low-rank structure of $\bMz$ implies a specific low-dimensional
nonlinear structure of $\bThetaz$.
For a common high missingness scenario, most entries of $\bMz$ are significantly
negative, where many common choices of the inverse link function can be
well-approximated by a linear function. So our modeling can be
regarded
as a low-rank modeling of $\bThetaz$.

There are some related but more specialized models.
\citet{Srebro-Salakhutdinov10} and \citet{Negahban-Wainwright12} utilize an independent row and column sampling mechanism, leading to a rank-1 structure for $\bThetaz$. \citet{Cai-Cai-Zhang16}
consider a matrix block structure for $\bThetaz$ and hence $\bMz$,
which can be regarded as a special case of the low-rank modeling.
\citet{Mao-Chen-Wong19} considered the case when the missingness is depends on observable covariates, and adopted a low-rank modeling with a known row space of $\bMz$.
The proposal in this paper is for the situation when the missingness is dependent on some hidden factors, which reflects situations when obvious covariates are unknown or not available.

\subsection{Inverse Probability Weighting Based Matrix Completion: Motivations and Challenges}
Write the Hadamard product as $\circ$ and
the Frobenius norm as $\norm{\cdot}_F$.
To recover the target matrix $\bAz$, many existing matrix
completion techniques
assume uniform missing structure
and hence utilize an unweighted/uniform empirical risk function
$\widehat{R}_{\text{UNI}}(\bA) = (n_1n_2)^{-1} \norm{\bW\circ(\bA-\bY)}_{F}^2$
\citep[e.g.,][]{Candes-Plan10,Koltchinskii-Lounici-Tsybakov11,Mazumder-Hastie-Tibshirani10},
which is an unbiased estimator of the risk 
$R(\bA)=\E(\|\bA-\bY\|^2_F)/(n_1 n_2)$ (up to a multiplicative constant) under uniform missingness.
The work of \citet{Klopp14} is a notable exception that considers the use of $\widehat{R}_{\text{UNI}}$
under non-uniform missingness. 

For any matrix
$\bB=(b_{ij})_{i,j=1}^{n_1,n_2}$, we {denote}
$\bB^{\dagger}=(b_{ij}^{-1})_{i,j=1}^{n_1,n_2}$ and $\bB^{\ddagger}=(b_{ij}^{-1/2})_{i,j=1}^{n_1,n_2}$.
Under general missingness (uniform or non-uniform),
one can show that,
for any $\bA\in\mbR^{n_1\times n_2}$,
\[
R\left(\bA\right)
=\frac{1}{n_1n_2}\E\left(\Norm{\bA-\bY}_{F}^2\right)
=\frac{1}{n_1n_2}\E\left(\Norm{\bW\circ\bhinvThetaz\circ\left(\bA-\bY\right)}_{F}^2\right).
\]
Clearly, $\bAz$ uniquely minimizes $R$.
If $\bTheta_{\star}$ were known, an unbiased estimator of $R$ would be
\begin{equation}
\widehat{R}\left(\bA\right)
=\frac{1}{n_1n_2} \Norm{\bW\circ\bhinvThetaz\circ\left(\bA-\bY\right)}_{F}^2,
\label{eqn:Rhat}
\end{equation}
which motivates the use of inverse probability weighting in matrix completion as in \citet{Mao-Chen-Wong19}.
In addition, our theoretical analysis shows that
the nuclear-norm-regularized empirical risk estimator (to be defined in details later)
based on $\widehat{R}$ (assuming the use of true observation probabilities)
improves upon existing error upper bound of corresponding estimator
based on $\widehat{R}_{\text{UNI}}$
achieved by \citet{Klopp14} as shown in Section \ref{sec:comp}.
However, the inverse probability weights $\bhinvThetaz$
are often unknown and have to be
estimated, which has to be conducted carefully in the context of matrix completion.

Despite the popularity of inverse probability weighting in missing data literature,
it is known to produce
unstable estimation due to occurrences of small probabilities
\citep[e.g.,][]{Rubin01,Kang-Schafer07,Schafer-Kang08}.
This problematic scenario is common for
matrix completion problems for recovering a target matrix from
very few observations.
Theoretically, a reasonable setup should allow
some $\theta_{\star,ij}$
to go to zero as $n_1,n_2\rightarrow \infty$,
leading to diverging  
weights and
a non-standard setup of inverse probability weighting.
Due to these observations, a careful construction of the estimation procedure is required.

For uniform sampling ($\theta_{\star,ij} \equiv \theta_0$ for some probability $\theta_0$),
one only has to worry about
a small common probability $\theta_0$
(or that $\theta_{0}$ diminishes in an asymptotic sense.)
Although small $\theta_0$ increases the difficulty of estimation,
$\widehat{R}(\bA)$ changes only up to a multiplicative constant.
However, for non-uniform setting,
it is not as straightforward due to the {heterogeneity}
among $\{\theta_{\star, ij}\}$. 
To demonstrate the issue, 
we now briefly look at
the Yahoo!~Webscope dataset described in Section \ref{sec:real}.
A sign of 
the strong heterogeneity
in  $\{\theta_{\star, ij}\}$ is
a large $\theta_U/\theta_L$,
where 
$\theta_L=\min_{i,j}\theta_{\star,ij}$ and
$\theta_U=\max_{i,j}\theta_{\star,ij}$. 
We found that 
the corresponding ratio of estimated probabilities $\widehat{\theta}_U/\widehat{\theta}_L$ based on {the} rank-1 structure of \cite{Negahban-Wainwright12} {was} $25656.2$, and that based on our proposed method (without re-estimation, 
to be described below) was $23988.0$.
This 
strong heterogeneity can jeopardize the convergence rate
of our estimator, which
will be properly addressed in our framework.

In the following section, we propose an estimation approach for $\bThetaz$ in Section \ref{sec:mleest}
and an appropriate modification in Section \ref{sec:constrainedest} which, when substituted into {the empirical risk} $\widehat{R}$, allows us to construct a stable estimator for $\bAz$.

\setcounter{equation}{0} 

\section{Estimation of \texorpdfstring{$\bThetaz$}{Theta0}}\label{sec:estTheta}

\subsection{Regularized Maximum Likelihood Estimation}\label{sec:mleest}
We develop the estimation of $\bThetaz$ based upon the framework of regularized maximum likelihood.
Given the inverse of link function $f$, the log-likelihood function with respect to the indicator matrix
$\bW=(w_{ij})\in\mbR^{n_1\times n_2}$ is
\[
\ell_{\bW}\left(\bM\right) = \sum_{i,j}  \left[\mathds{I}_{\left[w_{ij} = 1\right]}\log\left\{f\left(m_{ij}\right)\right\} + \mathds{I}_{\left[w_{ij} = 0\right]} \log\left\{1 - f\left(m_{ij}\right)\right\}\right],
\]
for any $\bM = (m_{ij})_{i,j=1}^{n_1,n_2}\in \mbR^{n_1\times n_2}$,
where $\mathds{I}_\mathcal{A}$ is the indicator of an event $\mathcal{A}$.
Due to the low-rank assumption of $\bMz$, one natural candidate
of estimators to $\bMz$ is the maximizer of the regularized log-likelihood
$\ell_{\bW}(\bM) - \lambda \|\bM\|_{*}$,
where $\|\cdot\|_*$ represents the nuclear norm
and $\lambda>0$ is a tuning parameter.
It is also common to enforce an additional max-norm constraint
$\|\bM\|_\infty\le \alpha$ for some $\alpha>0$ in the maximization \citep[e.g.,][]{Davenport-Plan-Berg14}.
Note that the nuclear norm penalty flavors
$\bM=0$, corresponding to that $\Pr(w_{ij}=1)=0.5$ for all $i,j$.
Nevertheless, this would not align well with common settings of matrix completion
under which the average probability of observations is quit small,
and hence would result in a large bias. In view of this, we instead adopt a parametrization $\bMz=\muz\bJ+\bZz$ and consider the following
estimator of $(\muz,\bZz)$: 
\begin{align}\label{prob:minmuZ}
\left(\bhmu,\bhZ\right) = &\underset{(\mu,\bZ)\in \mathcal{C}_{n_1,n_2}\left(\alpha_1,\alpha_2\right)}{\arg\max} \ell_{\bW}\left(\mu\bJ+\bZ\right)-\lambda
\Norm{\bZ}_{\ast}, \hbox{where} 
\end{align}
\[
\mathcal{C}_{n_1,n_2}\left(\alpha_1,\alpha_2\right)=\{ \left(\mu,\bZ\right)\in\mbR\times\mbR^{n_1\times n_2}:
\Abs{\mu}\le\alpha_1,\,\Norm{\bZ}_{\infty} \le \alpha_2,\,
\bm{1}_{n_1}^{\T}\bZ\bm{1}_{n_2}=0
\}.
\]
Note that the mean $\mu$
of the linear predictor $\mu\bJ+\bZ$ is not penalized. The
constraint $\bm{1}_{n_1}^{\T}\bZ\bm{1}_{n_2}=0$ ensures
the identifiability of $\mu$ and $\bZ$.
Apparently, the constraints in $\mathcal{C}_{n_1,n_2}\left(\alpha_1,\alpha_2\right)$
are analogous to 
$\|\bM\|_{\infty}\le \alpha_0$, where $\alpha_{0}=\alpha_1+\alpha_2$, but on the parameters $\mu$ and $\bZ$ respectively. With $(\bhmu, \bhZ)$, the corresponding estimator of $\bMz$ is $\bhM=\bhmu\bJ+\bhZ$.

\citet{Davenport-Plan-Berg14} considered
a regularized maximum likelihood approach for
a binary matrix completion problem.
Their goal 
was different, as they aimed at recovering a binary rating matrix in lieu of the missing structure,
and considered a regularization on $\bM$ (instead of $\bZ$)
via
$\norm{\bM}_{\ast}\le
\alpha'\{\mathrm{rank}(\bMz)n_1n_2\}^{1/2}$.
As for the scaling parameter $\alpha'$,
\citet{Davenport-Plan-Berg14} considered an $\alpha'$ independent of
the dimensions $n_1$ and $n_2$ to restrict the ``spikiness" of $\bM$.
As explained earlier, in our framework,
$\theta_L$
should be allowed to go to zero as $n_1,n_2\rightarrow \infty$.
To this end, we allow
$\alpha_1$ and $\alpha_2$ to depend on the dimensions 
$n_1$ and $n_2$.
See more details in Section \ref{sec:theory}.

\subsection{Computational algorithm and tuning parameter selection}
To solve the optimization \eqref{prob:minmuZ}, we begin with the observation that $\ell_{\bW}$ is a smooth concave function, which allows the usage of an iterative algorithm called accelerated proximal gradient algorithm \citep{Beck-Teboulle09}.
Given a pair $(\mu_{\text{old}},\bZ_{\text{old}})$ 
from {a} previous iteration,
a quadratic approximation of the objective function $-\ell_{\bW}(\mu\bJ+\bZ)+\lambda \norm{\bZ}_{\ast}$ is formed:
\begin{align*}
P_{L}\left\{\left(\mu,\bZ\right),\left(\mu_{\text{old}},\bZ_{\text{old}}\right)\right\}=&-\ell_{\bW}\left(\mu_{\text{old}}\bJ+\bZ_{\text{old}}\right)\\
&+\left(\mu-\mu_{\text{old}}\right)\bm{1}_{n_1}^{\T}\left\{-\nabla_{\mu}\ell_{\bW}\left(\mu_{\text{old}}\bJ+\bZ_{\text{old}}\right)\right\}\bm{1}_{n_2}+\frac{Ln_1n_2}{2}\left(\mu-\mu_{\text{old}}\right)^2\\
&+\Inner{\bZ-\bZ_{\text{old}}}{-\nabla_{\bZ}\ell_{\bW}\left(\mu_{\text{old}}\bJ+\bZ_{\text{old}}\right)}+\frac{L}{2}\Norm{\bZ-\bZ_{\text{old}}}_{F}^2+\lambda \Norm{\bZ}_{\ast},
\end{align*}
where $L>0$ is an algorithmic parameter
determining the step size of the proximal gradient algorithm,
and is chosen by a backtracking method \citep{Beck-Teboulle09}.
Here $\langle \bB, \bC\rangle = \sum_{i,j} b_{ij}c_{ij}$
for any matrices $\bB=(b_{ij})$ and $\bC=(c_{ij})$ of same dimensions.

In this iterative algorithm, a successive update of $(\mu,\bZ)$ can be obtained by
\[
\underset{\left(\mu,\bZ\right)\in \mathcal{C}_{n_1,n_2}\left(\alpha_1,\alpha_2\right)}{\arg\min}\quad P_{L}\left\{\left(\mu,\bZ\right),\left(\mu_{\text{old}},\bZ_{\text{old}}\right)\right\},
\]
where the optimization with respect to $\mu$ and $\bZ$ can be performed separately.
For $\mu$,
one can derive a closed-form update
\[
\min\left[\alpha_1,\max\left[-\alpha_1,\mu_{\text{old}}+\left(Ln_1n_2\right)^{-1}\bm{1}_{n_1}^{\T}\left\{-\nabla_{\mu}\ell_{\bW}\left(\mu_{\text{old}}\bJ+\bZ_{\text{old}}\right)\right\}\bm{1}_{n_2}\right]\right].
\]
As for $\bZ$, we need to perform the minimization 
\begin{align*}
&\underset{\Norm{\bZ}_{\infty} \le \alpha_2,\,
	\bm{1}_{n_1}^{\T}\bZ\bm{1}_{n_2}=0}{\arg\min}\quad \Inner{\bZ-\bZ_{\text{old}}}{-\nabla_{\bZ}\ell_{\bW}\left(\mu_{\text{old}}\bJ+\bZ_{\text{old}}\right)}+\frac{L}{2}\Norm{\bZ-\bZ_{\text{old}}}_{F}^2+\lambda \Norm{\bZ}_{\ast},
\end{align*}
which is equivalent to 
\begin{align}\label{prob:minZ}
&\underset{\Norm{\bZ}_{\infty} \le \alpha_2,\,
	\bm{1}_{n_1}^{\T}\bZ\bm{1}_{n_2}=0}{\arg\min}\quad \frac{1}{2}\Norm{\bZ-\bZ_{\text{old}}-\frac{1}{L}\nabla_{\bZ}\ell_{\bW}\left(\mu_{\text{old}}\bJ+\bZ_{\text{old}}\right)}_{F}^2+\frac{\lambda}{L}\Norm{\bZ}_{\ast}.
\end{align}
We apply a three-block extension of the alternative direction method of multipliers \citep{Chen-He-Ye16} to an equivalent form of \eqref{prob:minZ}:
\begin{equation}\label{prob:ADMMminZ}
\underset{\bZ=\bG_1=\bG_2,\,\bm{1}_{n_1}^{\T}\bG_1\bm{1}_{n_2}=0,\,\Norm{\bG_2}_{\infty} \le \alpha_2}{\arg\min}\quad \frac{\lambda}{L}\Norm{\bZ}_{\ast}+ \frac{1}{2}\Norm{\bG_2-\bZ_{\text{old}}-\frac{1}{L}\nabla_{\bZ}\ell_{\bW}\left(\mu_{\text{old}}\bJ+\bZ_{\text{old}}\right)}_{F}^2.
\end{equation}
Write $\bH=(\bH_1,\bH_2)$. The augmented
Lagrangian for \eqref{prob:ADMMminZ}
is
\begin{align*}
\mathcal{L}_{u}\left(\bZ,\bG_1,\bG_2;\bH\right)=&\frac{\lambda}{L}\Norm{\bZ}_{\ast}+\frac{1}{2}\Norm{\bG_2-\bZ_{\text{old}}-\frac{1}{L}\nabla_{\bZ}\ell_{\bW}\left(\mu_{\text{old}}\bJ+\bZ_{\text{old}}\right)}_{F}^2\\
&-\Inner{\bH_1}{\bZ-\bG_1}-\Inner{\bH_2}{\bZ-\bG_2}+\frac{u}{2}\Norm{\bZ-\bG_1}_{F}^2+\frac{u}{2}\Norm{\bZ-\bG_2}_{F}^2\\
&+\mathds{I}_{\left[\bm{1}_{n_1}^{\T}\bG_1\bm{1}_{n_2}=0\right]}+\mathds{I}_{\left[\Norm{\bG_2}_{\infty} \le \alpha_2\right]},
\end{align*}
where $u>0$ is an algorithmic parameter and, $\mathds{I}_\mathcal{A}=0$ if the constraint $\mathcal{A}$ holds and $\infty$ otherwise.
The detailed algorithm to solve \eqref{prob:ADMMminZ} is summarized in Algorithm \ref{appalg:ADMM} in the supplementary material.
It is noted that, in general, the multi-block alternative direction method of multipliers may fail to converge for some $u>0$
\citep{Chen-He-Ye16}. In those cases, an appropriate selection of $u$ is crucial.
However, we are able to show that the form of our algorithm belongs to a special class \citep{Chen-He-Ye16} in which convergence is guaranteed for any $u>0$. Therefore, we simply set $u=1$. We summarize the corresponding convergence result in
the following theorem whose proof is provided in the supplementary material.

\begin{theorem}\label{thm:ADMM}
	The sequence $\{\bZ^{(k)},\bG_1^{(k)},\bG_2^{(k)}\}$, generated by Algorithm \ref{appalg:ADMM} in the supplementary material, converges to the global optimum of \eqref{prob:ADMMminZ}.
\end{theorem}

Notice that the alternative direction method of multipliers algorithm is nested within the proximal gradient algorithm.
But, from our practical experiences, both the number of inner iterations (alternative direction method of multipliers) and
outer iterations (proximal gradient) are
small, usually less than twenty in our numerical experiments. Similarly, we summarize the corresponding convergence result of the overall proximal gradient algorithm in the following theorem whose proof is provided in the supplementary material.

\begin{theorem}\label{thm:proximal}
	The estimator $(\bhmu,\bhZ)$ generated by the proximal gradient algorithm, converges to the global optimum of \eqref{prob:minmuZ}.
\end{theorem}

The tuning parameters $\alpha_1$ and $\alpha_2$ can be chosen according to prior knowledge of the problem setup, if available.
When a prior knowledge is not available, one can choose large values for these parameters.
Once these parameters are large enough,
our method is not sensitive to their specific values.
A more principled way to tune $\alpha_1$ and $\alpha_2$ is
a challenging problem and beyond the scope of this work.
As for $\lambda$, we adopt Akaike information criterion ($\AIC$)
where the degree of freedom is approximated by $r_{\bhM}(n_1+n_2-r_{\bhM})$.

\subsection{Constrained estimation}\label{sec:constrainedest}
To use $\widehat{R}$ of (\ref{eqn:Rhat}), a naive idea is to obtain
$\bhTheta = (\hat{\theta}_{ij})_{i,j=1}^{n_1,n_2} = \mathcal{F}({\widehat{\bM}})$, 
where $\mathcal{F}(\bM)=
(f(m_{ij}))_{i,j=1}^{n_1,n_2}\in\mbR^{n_1\times n_2}$ 
for any $\bM=(m_{ij})_{i,j=1}^{n_1,n_2}\in\mbR^{n_1\times n_2}$,
and then replace $\bhinvThetaz$ by $\hatbhinvTheta=(\hat{\theta}_{ij}^{-1/2})_{i,j=1}^{n_1,n_2}$.
However, this direct implementation is not robust to 
extremely small probabilities of observation, 
and our theoretical analysis shows that this could lead to
a slower convergence rate in the estimation of $\bAz$.
In the literature of missing data,
a simple solution to robustify is winsorizing the small probabilities
\citep{Potter90,Scharfstein-Rotnitzky-Robins99}. 

In the estimation of $\bhTheta$ defined in (\ref{prob:minmuZ}) that assumes $\|\bZz\|_\infty\le \alpha_2$, a large $\alpha_2$ has an adverse effect on the estimation.
In the setting of diverging $\alpha_2$ (due to diminishing $\theta_L$),
the convergence rate of $\bhZ$ becomes slower and the estimator obtained
after direct winsorization will also be affected.
That is, even though the extreme probabilities could be controlled by winsorizing,
the unchanged entries of $\bhZ$ (in the procedure of winsorizing) may already suffer from a slower rate of convergence.
This results in
a larger estimation error under certain settings of missingness, as revealed in Section \ref{sec:theory}.

A seemingly better strategy is to impose a
tighter constraint directly in the minimization problem (\ref{prob:minmuZ}).
That is to adopt the constraint
$\norm{\bZ}_{\infty}\le\beta$
where $0\le \beta \le \alpha_2$.
Theoretically, one can better control the
errors on those entries of magnitude smaller than $\beta$.
However,
the mean-zero constraint of $\bZ$ no longer makes sense
as the constraint $\|\bZ\|_\infty\le \beta$ may have shifted
the mean.

We propose a re-estimation of $\bZz$
with a different constraint level $\beta$:
\begin{align}\label{prob:minZt}
\bhZ_{\beta} = \underset{\bZ\in \mbR^{n_1\times n_2}}{\arg\max} \ell_{\bW}\left(\bhmu\bJ+\bZ\right)-\lambda^{\prime} \Norm{\bZ}_{\ast}
\quad
\mathrm{subject~to} \quad \Norm{\bZ}_{\infty} \le \beta.
\end{align}
Note that we only re-compute $\bZ$ but not $\mu$, which allows us to drop the mean-zero constraint. Thus, $\bhM_{\beta}=\bhmu\bJ+\bhZ_{\beta}$.
The corresponding algorithm for optimization \eqref{prob:minZt}
can be derived similarly as in \citet{Davenport-Plan-Berg14},
and is provided in the supplementary material.
In what follows, we write $\bhTheta=\mcF(\bhM)$ and $\bhTheta_{\beta}=\mcF(\bhM_{\beta})$.

\setcounter{equation}{0} 

\section{Estimation of \texorpdfstring{$\bAz$}{A0}}\label{sec:estA0}
Now, we come back to \eqref{eqn:Rhat} and replace $\bhinvThetaz$ by 
$\hatbhinvTheta_{\beta}$ to obtain
a modified empirical risk:
\begin{equation}\label{eqn:risk}
\widetilde{R}\left(\bA\right)
=\frac{1}{n_1n_2} \Norm{\bW\circ\hatbhinvTheta_{\beta}\circ\left(\bA-\bY\right)}_{F}^2,
\end{equation}
where 
$\hatbhinvTheta_{\beta}=(\htheta_{ij,\beta}^{-1/2}) \in \mbR^{n_1\times n_2}$.
Since $\bA$ is a high-dimensional parameter,
a direct minimization of $\hat{R}^{\ast}$ often results in over-fitting.
To circumvent it,
we consider a regularized version:
\begin{equation}\label{eqn:objfun}
\widetilde{R}\left(\bA\right)+\tau\Norm{\bA}_{\ast},
\end{equation}
where
$\tau>0$ is a regularization parameter.
Again, the nuclear norm regularization
encourages low-rank solution.
Based on \eqref{eqn:objfun},
our estimator of $\bA_\star$ is 
\begin{equation}\label{prob:minA}
\bhA_{\beta}=\underset{\Norm{\bA}_{\infty}\le a}{\arg\min}\left\{\frac{1}{n_1n_2}\Norm{\bW\circ\hatbhinvTheta_{\beta}\circ\left(\bA-\bY\right)}^{2}_{F}+\tau\Norm{\bA}_{\ast}\right\},
\end{equation}
where $a$ is an upper bound on $\norm{\bAz}_{\infty}$. The above $\bhA_{\beta}$ contains as special cases (i) the matrix completion $\bhA_{\alpha_2}$, with unconstrained probability estimator $\bhTheta$, by setting 
$\beta=\alpha_2$ and (ii) the estimator $\bhA_{\beta}$, with constrained probability estimator $\bhTheta_{\beta}$, when $\beta < \alpha_2$.

We 
use an accelerated proximal gradient algorithm \citep{Beck-Teboulle09} to
solve \eqref{prob:minA}.
For the choice of tuning parameter $\tau$ in
\eqref{prob:minA}, we adopt a $5$-fold cross-validation
to select the remaining tuning parameters.
Due to the non-uniform missing mechanism, we
use a weighted version of the validation errors. The
specific details are given in Algorithm \ref{appalg:A} in the supplementary material.

\setcounter{equation}{0} 

\section{Theoretical Properties}\label{sec:theory}

\subsection{Probabilities of observation}\label{sec:asympTheta}
Let $\norm{\bB}=\sigma_{\max}(\bB)$, $\norm{\bB}_{\infty}=\max_{i,j}\vert
b_{ij}\vert$ and $\norm{\bB}_{\infty,2}=(\max_{i}\sum_{j}b_{ij}^{2})^{1/2}$ be
the spectral norm, the maximum norm and $l_{\infty,2}$-norm of a matrix $\bB$
respectively. We use the symbol $\asymp$ to represent the asymptotic
equivalence in order, i.e., $a_n\asymp b_n$ if $a_n=O(b_n)$ and
$b_n=O(a_n)$.
The average squared distance between two matrices $\bB, \bC\in \mathbb{R}^{n_1\times n_2}$
is $d^2(\bB, \bC) = \norm{\bB-\bC}_F^2/(n_1 n_2)$.
The average squared
errors of $\bhM_{\beta}$ and $\hatbinvTheta_{\beta}$ are then
$d^{2}(\bhM_{\beta},\bMz)$
and $d^{2}(\hatbinvTheta_{\beta},\binvThetaz)$ respectively.
We adopt the
Hellinger distance for any two matrices $S, T \in [0,1]^{n_1 \times n_2}$,
$d_{H}^2(S, T) = (n_1 n_2)^{-1} \sum_{i,j=1}^{n_1,n_2} d_{H}^2(s_{ij},
t_{ij})$ where
$d_{H}^2(s,t)=(s^{1/2}-t^{1/2})^2+\{(1-s)^{1/2}-(1-t)^{1/2}\}^2$
for $s,t \in [0,1]$.
In the literature of matrix completion,
most discussions related to optimal convergence rate
are only up to certain polynomial orders of $\log n$.
For convenience, we use $\polylog(n)$ for polynomials of $\log n$.

To investigate the asymptotic properties of $\bhM_{\beta}$ and $\hatbinvTheta_{\beta}$ defined in Section \ref{sec:estTheta}, we introduce the following conditions on the missing structure.

C1. The indicators
$\{w_{ij}\}_{i,j=1}^{n_1,n_2}$ are mutually independent,
and independent of $\{\epsilon_{ij}\}_{i,j=1}^{n_1,n_2}$.
For $i=1,\dots,n_1$ and $j=1,\dots,n_2$, $w_{ij}$
follows a Bernoulli distribution with probability of success 
$\theta_{\star,ij}=f(m_{\star,ij})\in (0,1)$.
Furthermore, $f$ is monotonic increasing and differentiable.

C2. The hidden feature matrix $\bMz=\muz\bJ+\bZz$ where $\bm{1}_{n_1}^T\bZz\bm{1}_{n_2}=0$,  $\abs{\muz}\le\alpha_1<\infty$ and $\norm{\bZz}_{\infty}\le\alpha_2<\infty$. Here $\alpha_1$ and $\alpha_2$ are allowed to depend on the dimensions $n_1$ and $n_2$. This also implies that there exists a lower bound $\theta_{L}\in (0,1)$ (allowed to depend on $n_1,n_2$) such that $\underset{i,j}{\min}\{\theta_{ij}\}\ge\theta_{L}\ge f(-\alpha_1-\alpha_2)>0$. 

For the convenience of the theoretical analysis, we consider an equivalent
estimator of $(\muz,\bZz)$ defined by the constrained maximization problem \eqref{prob:minmuZcon} instead of the Lagrangian form \eqref{prob:minmuZ}. For $r_{\bZz}\le\min\{n_1,n_2\}$ and $\alpha_1,\alpha_2\ge0$, let
\begin{align}\label{prob:minmuZcon}
\left(\bhmu,\bhZ\right) = &\underset{(\mu,\bZ)\in \widetilde{\mathcal{C}}_{n_1,n_2}\left(\alpha_1,\alpha_2\right)}{\arg\max} \ell_{\bW}\left(\mu\bJ+\bZ\right), \hbox{where} 
\end{align}
\begin{align*}
\widetilde{\mathcal{C}}_{n_1,n_2}\left(\alpha_1,\alpha_2\right)=&\left\{ \left(\mu,\bZ\right)\in\mbR\times\mbR^{n_1\times n_2}:
\Abs{\mu}\le\alpha_1,\Norm{\bZ}_{\infty} \le \alpha_2,\right.\nonumber\\
&\left.\Norm{\bZ}_{\ast}\le\alpha_2\sqrt{r_{\bZz}n_1n_2},
\bm{1}_{n_1}^{\T}\bZ\bm{1}_{n_2}=0
\right\}.
\end{align*}
It is easy to see that we have $(\mu_\star, \bZz)\in\widetilde{\mathcal{C}}_{n_1,n_2}(\alpha_1,\alpha_2)$ once $(\mu_\star, \bZz)\in\mathcal{C}_{n_1,n_2}(\alpha_1,\alpha_2)$ holds.
For the ease of presentation,
we assume $n_1=n_2=n$ and choose the logit function as the inverse link function $f$
in the rest of Section \ref{sec:theory}, while
corresponding results under general settings of $n_1$, $n_2$ and $f$ are delegated to Section \ref{appsec:gentheory} in the supplementary material. 
We first establish the convergence results for $\bhmu$, $\bhZ$ and $\bhM$, respectively. To simplify notations, let $\alpha_{0}=\alpha_1+\alpha_2$, $h_{\alpha_1,\beta}=(1+e^{\alpha_1+\beta})^{-1}$ and $\Gamma_{n}=e^{\alpha_0}(\alpha_1+\alpha_2r^{1/2}_{\bZz})n^{-1/2}$.

\begin{lemma}\label{lem:recover distribution (no constrain)}
	Suppose Conditions C1-C2 hold, and $(\mu_\star, \bZz)\in\mathcal{C}_{n_1,n_2}(\alpha_1,\alpha_2)$. Consider $\bhM=\bhmu\bJ+\bhZ$ where $(\bhmu,\bhZ)$ is the solution to~\eqref{prob:minmuZcon}.
	There exist {positive} constants $C_1, C_2$ such that we have
	with probability at least $1 - C_1/n$,
	\begin{align}\label{eqn:thm2}
	&\left(\muz-\bhmu\right)^2\le C_2 \left(\alpha_1^2\, \wedge \,\Gamma_{n}\right), \quad
	\frac{1}{n^2}\Norm{\bhZ -\bZz}_F^2\le C_2 \left(\alpha_2^2 \,\wedge\,\Gamma_{n}\right)\nonumber\\
	&\mbox{and}\quad
	\frac{1}{n^2}\Norm{\bhM -\bMz}_F^2\le C_2 \left(\alpha_0^2 \,\wedge\,\Gamma_{n}\right).
	\end{align}
\end{lemma}
The upper bounds in
\eqref{eqn:thm2}
all consist of trivial bounds $\alpha_j^2$
and a more dedicated bound
$\Gamma_{n}$.
The trivial upper bounds
$\alpha_1^2$, $\alpha_2^2$ and $\alpha_0^2$
can be easily derived from the constraint set $\mathcal{C}_{n_1,n_2}(\alpha_1,\alpha_2)$.
For extreme settings of increasing $\alpha_0$,
the more dedicated bound $\Gamma_{n}$ is diverging
and the trivial bounds may provide better control.
The term $\Gamma_{n}$ can be controlled by the rank of $\bZz$.
For a range of non-extreme scenarios, i.e., $\alpha_{0}\le 1/2\log n$ or $\theta_{L}\ge n^{-1/2}$,
the second term in $\Gamma_{n}$ attains the convergence order once $r_{\bZz}=O(1)$. 

Similarly, we study the theoretical results of the re-estimation of $\bZz$
  in terms of the constrained optimization:
	\begin{align}\label{prob:minZtcon}
	\bhZ_{\beta} = \underset{\bZ\in \mbR^{n_1\times n_2}}{\arg\max} \ell_{\bW}\left(\bhmu\bJ+\bZ\right)\nonumber
	\\
	\mathrm{subject~to} \quad \Norm{\bZ}_{\infty} \le \beta,\quad\Norm{\bZ}_{\ast}\le\beta\sqrt{r_{\mcT_{\beta}(\bZz)}n_1n_2},.
	\end{align}
We now consider the constrained estimation for $\bZz$, $\bMz$ and $\binvThetaz$.
For any matrix $\bB=(b_{ij})_{i,j=1}^{n_1,n_2}$, define the winsorizing operator $\mcT_{\beta}$ by $\mcT_{\beta}(\bB)=(T_{\beta}(b_{ij}))$ where
\begin{equation}\label{eq:T-beta}
T_{\beta}(b_{ij})=b_{ij}\mathds{I}_{[-\beta\le b_{ij}\le \beta]}+\beta\mathds{I}_{[b_{ij}>\beta]}-\beta\mathds{I}_{[b_{ij}<-\beta]}\quad\text{for any }\beta\ge 0. 
\end{equation} 
Write $\bM_{\star,\beta}=\muz\bJ+\mcT_{\beta}(\bZz)$ and $\bhM_{\star,\beta}=\bhmu\bJ+\mcT_{\beta}(\bZz)$, 
and $\bTheta_{\star,\beta}=\mcF(\bM_{\star,\beta})$ and $\bhTheta_{\star,\beta}=\mcF(\bhM_{\star,\beta})$ respectively.  
It is noted that  $\bhM_{\star,\beta}$ serves as a ``bridge" between the underlying  $\bM_{\star,\beta} $ and the empirical $\bhM_{\beta}$. 
Write
$N_{\beta}=\sum_{i,j}(\mathds{I}_{[z_{\star,ij}>\beta]}+\mathds{I}_{[z_{\star,ij}<-\beta]})$ as the number of extreme values in $\bZz$ at level $\beta$. The convergence rates of $d^{2}(\bhM_{\beta},\bMz)$ and $d^{2}(\hatbinvTheta_{\beta},\binvThetaz)$
are investigated in the next theorem. Define $\Lambda_{n}=\min[\beta^2,\tGamma_{n}+h_{\alpha_1,\beta}^{-1}n^{-2}\beta\{8N_{\beta}+(n^2-N_{\beta})\abs{\muz-\bhmu}\}]$ where $\tGamma_{n}=h_{\alpha_1,\beta}^{-1}(\alpha_1+\beta r^{1/2}_{\mcT_{\beta}(\bZz)})n^{-1/2}$.
\begin{theorem}
	\label{thm:recover M theta}
	Assume that Conditions C1-C2 hold, and $(\mu_\star, \bZz)\in\widetilde{\mathcal{C}}_{n_1,n_2}(\alpha_1,\alpha_2)$.
	Consider $\bhM_{\beta}=\bhmu\bJ+\bhZ_{\beta}$
	where $\bhZ_{\beta}$ is the solution to~\eqref{prob:minZtcon} and $\beta\ge0$, there exist some positive constants $C_1$, $C_2$ and $C_3$ such that we have with probability at least $1 - 2C_1/n$,
	\begin{equation*}\label{eqn:tight bound on M}
	d^{2}\left\{\bhZ_{\beta},\mcT_{\beta}\left(\bZz\right)\right\} \le C_3\Lambda_{n},\quad d^{2}\left(\bhM_{\beta},\bMz\right) \le C_2 \left(\alpha_1^2\, \wedge \,\Gamma_{n}\right)+C_3\Lambda_{n}+\frac{2(\alpha_2-\beta)_{+}^2N_{\beta}}{n^2}
	\end{equation*} 
	\begin{equation}\label{eqn:tight bound on invTheta}
	\quad\mbox{and}\quad d^{2}\left(\hatbinvTheta_{\beta},\binvThetaz\right)\le\frac{C_2}{h_{\alpha_1,\beta}^2}\left(\alpha_1^2\, \wedge \,\Gamma_{n}\right)
	+\frac{C_3\Lambda_{n}}{h_{\alpha_1,\beta}^2}+\frac{8N_{\beta}}{n^2\theta^2_{L}}.
	\end{equation}
\end{theorem}

We can derive an upper bound $4\beta^{2}$ for $d^{2}(\bhZ_{\textsf{Win},\beta},\mcT_{\beta}(\bZz))$ from the second term in Theorem \ref{lem:recover distribution (no constrain)} where $\bhZ_{\textsf{Win},\beta}=\mcT_{\beta}(\bhZ)$ is directly winsorized from $\bhZ$. Obviously, the order of this upper bound is larger than or equal to $\Lambda_{n}$.
Moreover,
there are scenarios where
$\Lambda_{n}$ is a smaller order of $\beta^2$.
To illustrate, assume that both $\alpha_1 \asymp 1$ and $\beta\asymp 1$, we have $h_{\alpha_1,\beta}\asymp 1$. Once we have $N_{\beta}=o(n)$, $r_{\mcT_{\beta}(\bZz)}=o(n)$ and $\abs{\bhmu-\muz}=o(1)$, then $\Lambda_{n}=o(\beta^2)$.

With a more dedicated investigation of \eqref{eqn:tight bound on invTheta}, one
can derive an upper bound for
$d^{2}(\hatbinvTheta_{\beta},\hatbinvTheta_{\star,\beta})$,
which will be used in Section \ref{sec:asympA0}.
Denote $k_{\alpha_1,\alpha_2,n}^{\prime}=\min\{\alpha_1^{2},e^{\alpha_0}(\alpha_1+\alpha_2r^{1/2}_{\bZz})n^{-1/2}\}$,
such an upper bound is of order
$k_{\alpha_1,\alpha_2,\beta,n}h_{\alpha_1,\beta}^{-2}$ where 
\begin{align*}
k_{\alpha_1,\alpha_2,\beta,n}\asymp\min\left[\beta^2,h_{\alpha_1,\beta}^{-1}\beta \left\{8N_{\beta}+\left(n^2-N_{\beta}\right)k_{\alpha_1,\alpha_2,n}^{\prime 1/2}\right\}n^{-2}+h_{\alpha_1,\beta}^{-1}n^{-1/2}(\alpha_1+\beta r^{1/2}_{\mcT_{\beta}(\bZz)})\right].
\end{align*}

\subsection{Target matrix}\label{sec:asympA0} 
To study the convergence of $d^{2}(\bhA_{\beta},\bAz)$, we require the following conditions regarding the random errors $\bepsilon$ and the target matrix $\bAz$. Recall that $\bhA_{\beta}$ includes both the estimations obtained with the unconstrained estimator $\bhTheta$ and the constrained estimator $\bhTheta_{\beta}$ as $\bhA(\bhTheta)=\bhA_{\alpha_2}$ with $\beta=\alpha_2$.

C3. (a) The random errors $\{\epsilon_{ij}\}$ in Model (\ref{eqn:datmod}) are independently distributed random variables
such that $\E(\epsilon_{ij})=0$ and $\E(\epsilon^2_{ij})=\sigma_{ij}^2<\infty$ for all $i,j$. 
(b) For some finite positive constants
$c_{\sigma}$ and $\eta$, 
$\underset{i,j}{\max}\,\E|\epsilon_{ij}|^{l}\le\frac{1}{2}l!c_{\sigma}^{2}\eta^{l-2}$
for any positive integer $l\ge 2$.

C4. There exists a positive constant $a_0$ such that $\norm{\bAz}_{\infty}\le a_0$.

Denote 
$h_{(1),\beta}=\underset{i,j}{\max}(\theta_{\star,ij}^{-1}\theta_{\star,ij,\beta})$ and 
\begin{equation}\label{eqn:Deltang}
\Delta =\max\left\{\frac{\left(c_{\sigma}\vee a\right)e^{-\muz/2+\alpha_2-\beta+\Abs{\alpha_2/2-\beta}}\left(n\log n\right)^{1/2}}{n^2},\frac{\eta e^{\muz/2+\alpha_1+\Abs{\alpha_2/2-\beta}} k_{\alpha_1,\alpha_2,\beta,n}^{1/2}\log^{3/2}n}{h_{\alpha_1,\beta}n}\right\}.
\end{equation}
The following theorem established a general upper bound for $d^{2}(\bhA_{\beta},\bAz)$.

\begin{theorem}\label{thm:keythm}
	Assume Conditions C1-C4 hold.
	For $\beta\ge0$,  there exist some positive constants $C_4$, $C_5$, $C_6$ and $C_7$, both independent of $\beta$, such that for $h_{(1), \beta}\tau\ge C_4\Delta$, 
	we have with probability at least $1-3/(2n)$,
	\begin{equation}\label{eqn:keythm}
	d^{2}\left(\bhA_{\beta},\bAz\right)
	\le \max\left\{C_6n^2h_{\left(1\right),\beta}^2r_{\bAz}\tau^2+C_7h_{\left(1\right),\beta}^2h_{\left(2\right),\beta}^2r_{\bAz}n^{-1}\log\left(n\right),C_5h_{\left(1\right),\beta}h_{\left(3\right),\beta}n^{-1}\log^{1/2}(n)\right\}.
	\end{equation}
\end{theorem}

As for the estimator of the target matrix
based on direct winsorization
$\bhTheta_{\textsf{Win},\beta}=\mcF(\bhmu\bJ+\bhZ_{\textsf{Win},\beta})$ {where $\bhZ_{\textsf{Win},\beta}=\mcT_{\beta}(\bhZ)$},
an upper bound can be derived
using Theorem \ref{thm:recover M theta}.
As noted in a remark after Theorem \ref{thm:recover M theta},   $d^{2}(\bhZ_{\textsf{Win},\beta},\mcT_{\beta}(\bZz))$ converges at a slower rate $\beta^2$ which will cause 
a larger error bound for the target matrix. 

Now, we discuss the rates of $d^{2}(\bhA_{\beta},\bAz)$  
under various missing structures.
For simplicity, the following discussion focuses on the low-rank linear predictor   
($\bMz$) setting such that $r_{\bMz}\asymp 1$.

Uniform missingness. Under the uniform missingness, i.e., $\theta_{ij}\equiv\theta_{0}$, it has been
shown in \citet{Koltchinskii-Lounici-Tsybakov11} that $\theta_{0}^{-1}n^{-1}\polylog(n)$ is the optimal rate
for $d^{2}(\bhA_{\beta},\bAz)$.
Therefore it is reasonable to require
$\alpha_1+\alpha_2=\alpha_{0}=O(\polylog(n))$
for the convergence of $d^{2}(\bhA_{\beta},\bAz)$.
Under the uniform missingness, we have $\alpha_2=0$, $\alpha_0=\alpha_1$ and $e^{\muz}\asymp\theta_0$. For $\beta=0$, our estimator $\bhA_{\beta}$ 
degenerates to the estimator based on the unweighted empirical risk function. Theorem \ref{thm:keythm} shows that 
$\bhA_{\beta}$ 
achieves the optimal rate $\theta_{0}^{-1}n^{-1}\polylog(n)$.
As for $\beta>0$, by taking $\beta\to 0$ such that $k_{\alpha_1,\alpha_2,\beta,n}=O(e^{\muz-2\alpha_1-2\beta}n^{-1}\log^{-2}n)$, the estimator can also reach the optimal rate.
Of interest here is that $\beta$ is allowed to be strictly positive to achieve the same rate.

Non-uniform missingness. Under the non-uniform missingness, suppose the lower and upper bounds of observation probability satisfy
$\theta_L \asymp e^{\muz-\alpha_2}$ and $\theta_{U} \asymp e^{\muz+\alpha_2}$. For the non-constrained case 
of $\beta=\alpha_2$ and $h_{\alpha_1,\beta}\asymp e^{-\alpha_1-\alpha_2}$, 
the second term of $\Delta$  in \eqref{eqn:Deltang} dominates due to the
fact that 
$$e^{-\muz/2+\alpha_2/2}n^{-3/2}\log^{1/2} n=o(e^{\muz/2+5\alpha_1/2+3\alpha_2/2}n^{-5/4}\log^{3/2} n).$$  
Thus, the convergence rate of $d^{2}(\bhA_{\beta},\bAz)$ is $e^{\muz+5\alpha_1+3\alpha_2}n^{-1/2}\log^{3} n$.
To guarantee convergence, as $e^{\muz/2+5\alpha_1/2+3\alpha_2/2}\le e^{3\alpha_1+3\alpha_2/2}$, it requires that $\alpha_1+\alpha_2/2< (1/12)\log n$ which implies that $\theta_{L}^{-1}=O(n^{1/6})$. 

However, the above range of $\theta_L = O(n^{-1/6})$ excludes
$\theta_L\equiv(n^{-1}\polylog(n))$, the case that results in the number of the observed matrix entries at the order of $n\,\polylog(n)$ which represents
the most sparse case of observation where the matrix can still be recovered \citep{Candes-Recht09,Candes-Plan10,Koltchinskii-Lounici-Tsybakov11,Negahban-Wainwright12}.  
We will show in the following that  with an appropriately chosen $\beta$, 
the constrained estimator 
$\bhTheta_{\beta}$ can accommodate the case of $\theta_{L}^{-1}=O(n\log^{-1}n)$.

Case (I): $\beta=0$. To demonstrate this, we start with the absolute constrained case, i.e., $\beta=0$, 
which forces the estimated probabilities to be uniform and 
implies $e^{-\muz/2+\alpha_2-\beta+\abs{\alpha_2/2-\beta}}=e^{-\muz/2+3\alpha_2/2}\asymp\theta_{U}^{1/2}\theta_{L}^{-1}$. Then, according to Theorem \ref{thm:keythm}, $d^{2}(\bhA_{\beta},\bAz)$ attains the convergence rate $\theta_{U}\theta_{L}^{-2}n^{-1}\log (n)$, which converges to 0 provided $\theta_{U}\theta_{L}^{-2}=o(n\log^{-1}n)$.  
Obviously, the condition $\theta_{U}\theta_{L}^{-2}=o(n\log^{-1}n)$ includes the extreme case of $\theta_{L}^{-1}=O(n\log^{-1}n)$ and $n\,\polylog(n)$
observations.

Case (II): $\beta>0$. For the more interesting setting $\beta>0$, 
to simplify the discussion, we concentrate on the case when the first term in $k_{\alpha_1,\alpha_2,\beta,n}$
is of a smaller order,
which can be achieved by choosing $\beta=O(e^{-\muz-2\alpha_1+\alpha_2}n^{-1/2}\log^{-1} n)$.
Then, according to Theorem \ref{thm:keythm},
$$d^{2}(\bhA_{\beta},\bAz)=
O_p(e^{-\muz+2\alpha_2-2\beta+2\abs{\alpha_2/2-\beta}}n^{-1}\log n)=O_p(e^{\alpha_1/2+3\alpha_2/2}n^{-1}\log n),$$ 
since $e^{-\muz/2+\alpha_2-\beta+\abs{\alpha_2/2-\beta}}\le e^{\alpha_1/2+3\alpha_2/2}$. In the following we consider two further cases: 
(i) $\alpha_2=O((\log\log n)^{-1}\alpha_1)$
and 
(ii) $\alpha_1=o(\alpha_2\log\log n)$.
Note that for either cases, $e^{-\muz+2\alpha_2-2\beta+2\abs{\alpha_2/2-\beta}}\asymp \theta_{U}\theta_{L}^{-2}$ which leads to 
$$d^{2}(\bhA_{\beta},\bAz)=
O_p(\theta_{U}\theta_{L}^{-2}n^{-1}\log n).$$ 

If $\alpha_2=O\{(\log\log n)^{-1}\alpha_1\}$,  we require $\alpha_1<(1+3\log\log n)^{-1}(\log n-\log\log n)$ to guarantee
convergence, which implies that $\theta_{L}=O(n^{-1})$. Thus,  we only lose a  $\polylog(n)$  factor when compared with
the most extreme but feasible setting of $\theta_L^{-1}=O[n\{\polylog(n)\}^{-1}]$.  Also $\beta=O(e^{-\muz-2\alpha_1+\alpha_2}n^{-1/2}\log^{-1} n)$ implies that $\beta=O(n^{-1/2}\log^{-1} n)$. 

If $\alpha_1=o\{(\log\log n)\alpha_2\}$, we require that $\alpha_2< \{3+(\log\log n)^{-1}\}^{-1}(\log n-\log\log n)$ which leads to $\theta_{L}^{-1}=O(n^{1/3})$.
Also $\beta=O(e^{-\muz-2\alpha_1+\alpha_2}n^{-1/2}\log^{-1} n)$ implies that $\beta=O(n^{-1/6}\log^{-1}n)$. However, to make $d^{2}(\bhA_{\beta},\bAz)$ convergent,
the attained rate for $\theta_L^{-1}$ has to be $O(n^{1/3})$, which excludes the most extreme heterogeneity case of $\theta_L^{-1}=O\{n(\polylog(n))^{-1}\}$.  
The reason for not being able to cover the most extreme case of $\theta_L^{-1}=O\{n(\polylog(n))^{-1}\}$ is that the current Case (ii) allows more heterogeneity in $\bZz$ as reflected by 
having a larger $\alpha_2$ than that prescribed under Case (i).
As $\muz$ is jointly estimated with $\bZz$ in the unconstrained estimation
(Section \ref{sec:mleest}),
stronger heterogeneity slows down the convergence rate in the estimation of $\muz$,
which becomes a bottleneck for further improvement.
If $\muz$ was observable, the problem would not be as serious
despite the adverse effect of stronger heterogeneity on the estimation of $\bZz$.

To summarize, under the uniform missing and Case (I), (II)(i) in the non-uniform missing, we can achieve the optimal rate up to a $\polylog(n)$ order. For Case (II)(ii), when the missingness is not extreme, with an appropriately chosen $\beta>0$, the proposed estimator can also attain the optimal rate up to the $\polylog(n)$ order.

\subsection{Comparison with Uniform Objective Function}\label{sec:comp}
Recall that
the unweighted empirical risk function
$\widehat{R}_{\text{UNI}}(\bA) = n^{-2} \norm{\bW\circ(\bA-\bY)}_{F}^2$
is adopted by many existing matrix completion techniques \citep{Klopp14}.
An interesting question is whether there is any benefit
in adopting the proposed weighted empirical risk function for matrix completion. 
In this subsection,
we aim to shed some light on this aspect by
comparing the non-asymptotic error bounds
of the corresponding estimators.
Due to the additional complication from the estimation error of
the observation probability matrix,
we only focus on the weighted empirical risk function
with true inverse probability weighting in this section.
We will demonstrate empirically in Sections \ref{sec:sim} and \ref{sec:real}
the benefits of the weighted objective function with estimated weights.

Most existing work with unweighted empirical risk function assume the true missingness is uniform \citep{Candes-Plan10,Koltchinskii-Lounici-Tsybakov11}.
One notable exception is \citet{Klopp14}, where unweighted empirical risk function is
studied under possibly non-uniform missing structure.
The estimator of \citet{Klopp14} is equivalent to
our estimator when $\beta=0$, which is denoted by $\bhA^{\text{UNI}}$. Thus, according to Theorem \ref{thm:keythm}, we have with probability at least $1-3/(2n)$,
\[
d^{2}\left(\bhA^{\text{UNI}},\bAz\right)
\le \min\left\{(C_6+C_7)r_{\bAz}\theta_{U}\theta_{L}^{-2}n^{-1}\log n,C_5\theta_{U}^{1/2}\theta_{L}^{-1}n^{-1/2}\log^{1/2} n\right\}=U^{\text{UNI}},
\]
which is the same upper bound obtained in \citet{Klopp14}.
Define $\bhA^{\text{KNOWN}}$ as the estimator which minimizes the known weighted empirical risk function \eqref{eqn:Rhat}.
Then, 
\[
d^{2}\left(\bhA^{\text{KNOWN}},\bAz\right)
\le \min\left\{(C_6+C_7)r_{\bAz}\theta_{L}^{-1}n^{-1}\log n,C_5\theta_{L}^{-1/2}n^{-1/2}\log^{1/2} n
\right\}=U^{\text{KNOWN}}.
\]
The improvement in the upper bounds of the weighted objective function $\widehat{R}$ lies in that,
under non-uniform missingness, $\theta_{U}\theta_{L}^{-1}>1$ which implies that $U^{\text{KNOWN}} < U^{\text{UNI}}$ as summarized below. 

\begin{theorem}\label{thm:uniupperbound}
	Assume Conditions C1-C4 holds, and take $\tau_{\text{KNOWN}}=C_3\theta_{L}^{-1/2}n^{-3/2}\log^{1/2}n$ and $\tau_{\text{UNI}}= C_3\theta_{U}^{1/2}f^{-1}(\muz)n^{-3/2}\log^{1/2}n$.
	The upper bound of $d^{2}(\bhA^{\text{UNI}},\bAz)$ is the same as $U^{\text{UNI}}$ and the upper bound of $d^{2}(\bhA^{\text{KNOWN}},\bAz)$ is the same as $U^{\text{KNOWN}}$.
	In addition, $U^{\text{KNOWN}}\le U^{\text{UNI}}$, and $U^{\text{KNOWN}} < U^{\text{UNI}}$ if $\theta_{U} > \theta_{L}$, i.e., the true missing mechanism is non-uniform.
\end{theorem}

Our approach draws inspiration from the missing value literature, for instance in \citet{Chen-Leung-Qin08}, which showed that using the estimated parameters in the inverse probability weighting can actually reduce the variance of the parameter of interest; see Theorem 1 of the paper. Given the results of \citet{Chen-Leung-Qin08}, we would expect using the estimated parameters $\bhTheta_{\beta}$ in the weighting probability would not be inferior to the version with the true parameter $\bhTheta_{\star}$.

\setcounter{equation}{0} 

\section{Simulation Study}\label{sec:sim}
\subsection{Missingness}\label{sec:sim:miss}
This section reports results from simulation experiments which were designed to evaluate the numerical performance of the proposed methodologies.
We first evaluate the estimation performances of the observation probabilities
in Section \ref{sec:sim:miss}
and then those of the target matrix in Section \ref{sec:sim:target}.
In the simulation, the true observation probabilities $\bThetaz$ and the target
matrix $\bAz$ were randomly generated once and kept fixed for each simulation setting
to be described below. To generate $\bThetaz$, we
first generated $\bU_{\bMz}\in\mathbb{R}^{n_1\times (r_{\bMz}-1)}$ and
$\bV_{\bMz}\in\mathbb{R}^{(r_{\bMz}-1)\times n_2}$ as random Gaussian
matrices with independent entries each following $\mathcal{N}(-0.4,1)$.
We then obtained $\bMz=\bU_{\bMz}\bV_{\bMz}^{\T}-\bar{m}_{n_1,n_2,r_{\bMz}}\bJ$ where $\bar{m}_{n_1,n_2,r_{\bMz}}$ is a scalar chosen to ensure the average observation rate is $0.2$ in each simulation setting. We finally set $\bThetaz=\mcF(\bMz)$ where the inverse link function $f$ is a logistic function.

In our study, we set $r_{\bMz}=11$, (or $r_{\bZz}=10$) and chose $n_1=n_2$ with four sizes: $600$, $800$, $1000$ and $1200$, and the number of simulation runs for each settings was $500$.

For the purpose of benchmarking, we compared various estimators of the missingness:
\begin{enumerate}
	\item the non-constrained estimator  $\bhTheta_{\alpha}$ defined in \eqref{prob:minmuZ};
	\item the constrained estimator $\bhTheta_{\beta}$ defined in \eqref{prob:minZt};
	\item the directly winsorized estimator $\bhTheta_{\textsf{Win},\beta}=\mcF\{\bhmu\bJ+\mcT_{\beta}(\bhZ)\}$;
	\item the 1-bit estimator $\bhTheta_{\textsf{1-bit},\alpha}$ proposed in \citet{Davenport-Plan-Berg14} and its corresponding constrained and winsorized versions $\bhTheta_{\textsf{1-bit},\beta}$ and $\bhTheta_{\textsf{1-bit},\textsf{Win},\beta}$;
	(note that the 1-bit estimator $\bhTheta_{\textsf{1-bit},\alpha}$
	imposes the nuclear-norm regularization on the whole $\bM$ instead of $\bZ$, when compared to $\bhTheta_{\alpha}$)
	\item the rank-1 probability estimator $\bhTheta_{\textsf{NW}}$ used in \citet{Negahban-Wainwright12} where $g_{i.}=n_2^{-1}\sum_{j=1}^{n_2}w_{ij}$, $g_{.j}=n_1^{-1}\sum_{i=1}^{n_1}w_{ij}$ and $\theta_{ij,\textsf{NW}}=g_{i.}g_{.j}$;
	\item the uniform estimator $\bhTheta_{\textsf{UNI}}=N/(n_1n_2)\bJ$.
\end{enumerate}
For the non-constrained estimator $\bhTheta_{\alpha}$ and the 1-bit estimator $\bhTheta_{\textsf{1-bit},\alpha}$, the parameter $\alpha$ is set according to the knowledge of the true $\bMz$. For the constrained estimators $\bhTheta_{\beta}$ and $\bhTheta_{\textsf{Win},\beta}$, the constraint level $\beta$ was chosen so that either $5\%$ or $10\%$ of the elements in $\bhZ_{\alpha}$ were winsorized. Similarly for $\bhTheta_{\textsf{1-bit},\beta}$ and $\bhTheta_{\textsf{1-bit},\textsf{Win},\beta}$.

To quantify the estimation performance of linear predictor $\bMz$ and observation probabilities $\bThetaz$, we considered the empirical root mean squared errors $\text{RMSE}(\bB,\bC)$ with respect to any two matrices $\bB$ and $\bC$ of dimension $n_1\times n_2$,
and the Hellinger distance $d_{H}^{2}(\bhTheta,\bThetaz)$ between $\bhTheta$ and $\bThetaz$ defined as follows:
\[
\text{RMSE}\left(\bB,\bC\right)=\frac{\Norm{\bB-\bC}_{F}}{\left(n_1n_2\right)^{1/2}}
\quad\mbox{and}\quad
d_{H}^{2}\left(\bhTheta,\bThetaz\right)=\frac{\sum_{i,j=1}^{n_1,n_2}d_{H}^2\left(\htheta_{ij},\theta_{\star,ij}\right)}{\left(n_1n_2\right)^{1/2}}.
\]
As the estimators $\mathcal{F}^{-1}(\bhTheta_{\alpha})$ and $\mathcal{F}^{-1}(\bhTheta_{\textsf{1-bit},\alpha})$ are both low-rank, we also
report their corresponding ranks.

\linsps
\begin{table}[t!]
	\small
	\centering
	\caption{Root mean squared errors $\text{RMSE}(\bhM,\bMz)$, Hellinger distance $d_{H}^{2}(\bhTheta,\bThetaz)$, rank of linear predictor $\bhM$ and estimated $\bhTheta$ and their standard errors (in parentheses) under the low rank missing observation mechanism, with ($n_1,n_2$) = (600, 600), (800, 800), (1000, 1000), (1200, 1200) and $r_{\bMz}=11$, for the proposed estimators $\bhTheta_{\alpha}$, $\bhTheta_{\textsf{1-bit},\alpha}$ and the two existing estimators ($\bhTheta_{\textsf{NW}}$ and $\bhTheta_{\textsf{UNI}}$).}
	\label{tab:thetasimres}\par
	\begin{tabular}{rllll}
		\hline
		600 & $\bhTheta_{\alpha}$ & $\bhTheta_{\textsf{1-bit},\alpha}$ & $\bhTheta_{\textsf{NW}}$ & $\bhTheta_{\textsf{UNI}}$ \\
		\hline
		\text{RMSE}($\bhM,\bMz$) & 2.6923 (0.0342) & 2.9155 (0.0295) & - & - \\ 
		$d_{H}^{2}(\bhTheta,\bThetaz)$ & 0.0369 (0.0015) & 0.0450 (0.0016) & 0.1233 (1e-04) & 0.1729 (1e-04) \\
		$r_{\bhM}$ & 12.45 (0.50) & 12.69 (0.46) & - & - \\ 
		$r_{\bhTheta}$ & 600.00 (0.00) & 600.00 (0.00) & - & - \\
		\hline
		800 & $\bhTheta_{\alpha}$ & $\bhTheta_{\textsf{1-bit},\alpha}$ & $\bhTheta_{\textsf{NW}}$ & $\bhTheta_{\textsf{UNI}}$ \\
		\hline
		\text{RMSE}($\bhM,\bMz$) & 2.5739 (0.0116) & 2.7796 (0.0033) & - & - \\  
		$d_{H}^{2}(\bhTheta,\bThetaz)$ & 0.0317 (5e-04) & 0.0379 (1e-04) & 0.1219 (1e-04) & 0.1767 (1e-04) \\ 
		$r_{\bhM}$ & 12.04 (0.20) & 12.03 (0.17) & - & - \\ 
		$r_{\bhTheta}$ & 800.00 (0.00) & 800.00 (0.00) & - & - \\
		\hline
		1000 & $\bhTheta_{\alpha}$ & $\bhTheta_{\textsf{1-bit},\alpha}$ & $\bhTheta_{\textsf{NW}}$ & $\bhTheta_{\textsf{UNI}}$ \\
		\hline
		\text{RMSE}($\bhM,\bMz$) & 2.4870 (0.0212) & 2.7731 (0.0015) & - & - \\  
		$d_{H}^{2}(\bhTheta,\bThetaz)$ & 0.0266 (8e-04) & 0.0351 (1e-04) & 0.1246 (1e-04) & 0.1767 (1e-04) \\ 
		$r_{\bhM}$ & 12.68 (0.53) & 12.00 (0.00) & - & - \\ 
		$r_{\bhTheta}$ & 1000.00 (0.00) & 1000.00 (0.00) & - & - \\
		\hline
		1200 & $\bhTheta_{\alpha}$ &
		$\bhTheta_{\textsf{1-bit},\alpha}$ & $\bhTheta_{\textsf{NW}}$ & $\bhTheta_{\textsf{UNI}}$ \\
		\hline
		\text{RMSE}($\bhM,\bMz$) & 2.3809 (0.0018) & 2.6470 (0.0012) & - & - \\ 
		$d_{H}^{2}(\bhTheta,\bThetaz)$ & 0.0242 (1e-04) & 0.0314 (1e-04) & 0.1211 (1e-04) & 0.1761 (1e-04) \\ 
		$r_{\bhM}$ & 12.00 (0.00) & 12.00 (0.00) & - & - \\ 
		$r_{\bhTheta}$ & 1200.00 (0.00) & 1200.00 (0.00) & - & - \\ 
		\hline
	\end{tabular}
\end{table}
\linsp

Table \ref{tab:thetasimres} summarizes the simulation results for the
missingness. The most visible aspect of the results is that the proposed
estimators $\bhTheta_{\alpha}$ and $\bhTheta_{\textsf{1-bit},\alpha}$ both have
superior performance than the two existing estimators $\bhTheta_{\textsf{NW}}$ and $\bhTheta_{\textsf{UNI}}$ by having
smaller root mean square errors with respect to $\bhM$, Hellinger distances $d_{H}^{2}(\bhTheta,\bThetaz)$ and more
accuracy estimated rank of $\bMz$. Without the separation of $\muz$ from
$\bMz$, $\bhTheta_{\textsf{1-bit},\alpha}$ has larger error and Hellinger
distance than the proposed estimators. The performance of $\bhTheta_{\textsf{NW}}$
is roughly between the proposed estimators and the uniform estimator
$\bhTheta_{\textsf{UNI}}$. Estimator $\bhTheta_{\textsf{UNI}}$ is a
benchmark which captures no variation of the observation probabilities.

\subsection{Target matrix}\label{sec:sim:target}

To generate a target matrix $\bAz$, we first generated
$\bU_{\bAz}\in\mathbb{R}^{n_1\times (r_{\bAz}-1)}$ and
$\bV_{\bAz}\in\mathbb{R}^{(r_{\bAz}-1)\times n_2}$ as random matrices with
independent Gaussian entries distributed as $\mcN(0,\sigma_{\bAz}^2)$ and
obtained $\bAz=2.5\bJ+\bU_{\bAz}\bV_{\bAz}^{\T}$.
Here we set the standard deviation of the entries in the matrix product
$\bU_{\bAz}\bV_{\bAz}^{\T}$ to be $2.5$ to mimic the
Yahoo!~Webscope data set described in Section \ref{sec:real}.
To achieve this, $\sigma_{\bAz}=(2.5^{2}/(r_{\bAz}-1))^{1/4}$.
The contaminated version of $\bAz$ was then generated as
$\bY=\bAz+\bepsilon$, where $\bepsilon\in\mathbb{R}^{n_1\times n_2}$ has
i.i.d.~mean zero Gaussian entries
$\epsilon_{ij}\sim\mcN(0,\sigma_{\epsilon}^{2})$. The $\sigma_{\epsilon}^{2}$
is chosen such that
$\text{SNR}=(\E\norm{\bAz}_{F}^2/\E\norm{\bepsilon}_{F}^2)^{1/2}=1$, where
$\E\norm{\bAz}_{F}^2=n_1n_2(r_{\bAz}-1+2.5^2)$ implies
$\sigma_{\epsilon}=0.5(r_{\bAz}-1+2.5^2)^{1/2}$.

For the estimation of the target matrix, we evaluated ten versions of the
proposed estimators \textsf{Proposed\_$\bhTheta_{\beta}$\_$t$},
\textsf{Proposed\_$\bhTheta_{\textsf{Win},\beta}$\_$t$},
\textsf{Proposed\_$\bhTheta_{\alpha}$},
\textsf{Proposed\_$\bhTheta_{\textsf{1-bit},\beta}$\_$t$},
\textsf{Proposed\_$\bhTheta_{\textsf{1-bit},\textsf{Win},\beta}$\_$t$} and
\textsf{Proposed\_$\bhTheta_{\textsf{1-bit},\alpha}$}.
Here \textsf{Proposed}
indicates the estimators are obtained by solving problem \eqref{prob:minA},
while
$\bhTheta_{\beta}$, $\bhTheta_{\textsf{Win},\beta}$, $\bhTheta_{\alpha}$,
$\bhTheta_{\textsf{1-bit},\beta}$,
$\bhTheta_{\textsf{1-bit},\textsf{Win},\beta}$ and
$\bhTheta_{\textsf{1-bit},\alpha}$ represents the probability estimators used in
\eqref{prob:minA}, as described in Section \ref{sec:sim:miss}, and $t=0.05$ or $0.1$ denote the winsorized proportion for which $\beta$ is chosen.
In addition, same as \citet{Mao-Chen-Wong19}, we also compared them with three existing matrix completion techniques: the
methods proposed in \citet{Negahban-Wainwright12} (\textsf{NW}),
\cite{Koltchinskii-Lounici-Tsybakov11} (\textsf{KLT}) and
\cite{Mazumder-Hastie-Tibshirani10} (\textsf{MHT}).
Among these three methods, \textsf{NW} is the only one that
adjusts for non-uniform missingness.
All three methods require tuning parameter selection,
for which cross-validation is adopted.
See \citet{Mao-Chen-Wong19} for more details.

To quantify the performance of the matrix completion, in addition to the empirical root mean squared errors with respect to $\bhA_{\beta}$ and $\bAz$, we used one more measure:
\[
\text{Test Error}=\frac{\Norm{\bW^{\star}\circ\left(\bhA_{\beta}-\bAz\right)}_{F}^{2}}{\Norm{\bW^{\star}\circ\bAz}_{F}^{2}}, 
\]
where $\bW^{\star}$ is the matrix of missing indicator
with the $(i,j)$th
entry being $(1-w_{ij})$.
The test error measures the relative estimation error of the unobserved entries to their signal strength. The estimated ranks of $\bhA_{\beta}$ are also reported.

Tables \ref{tab:Asimresr11n1} summarize the simulation results
for different dimensions $n_1$=$n_2$ ranges from $600$ to $800$ and two
different settings of $r_{\bAz}=11$. The results of $r_{\bAz}=11$ for different dimensions $n_1$=$n_2$ ranges from $1000$ to $1200$ are delegated to Table \ref{apptab:Asimresr11n2} and the results of $r_{\bAz}=31$ are delegated to Tables \ref{apptab:Asimresr31n1}-\ref{apptab:Asimresr31n2} of Section \ref{appsec:sim} in the supplementary material.
From the tables, we notice
that the ten versions of the proposed methods possess
superior performance than the three existing methods by having smaller root mean squared errors and
Test Errors. Among the first five proposed methods in the tables,
\textsf{Proposed\_$\bhTheta_{\beta}$} is better than
\textsf{Proposed\_$\bhTheta_{\alpha}$} for most of the time. It is because that
the constrained estimator $\bhTheta_{\beta}$ has much smaller ratio
$\widehat{\theta}_{U}/\widehat{\theta}_{L}$ than $\bhTheta_{\alpha}$ which
improve the stability of prediction and the accuracy.
Another observation is that
\textsf{Proposed\_$\bhTheta_{\beta}$\_0.1} performs better than
\textsf{Proposed\_$\bhTheta_{\textsf{1-bit},\alpha}$} at most times.

\linsps
\begin{table}[t!]
	\small
	\caption{Root mean squared errors, test errors, estimated ranks $r_{\bhA_{\beta}}$ and their standard deviations (in parentheses) under the low rank missing observation mechanism, for three existing methods and ten versions of the proposed methods where \textsf{Proposed} indicates the estimators are obtained by solving problem \eqref{prob:minA}, while $\bhTheta_{\beta}$, $\bhTheta_{\textsf{Win},\beta}$, $\bhTheta_{\alpha}$, $\bhTheta_{\textsf{1-bit},\beta}$,$\bhTheta_{\textsf{1-bit},\textsf{Win},\beta}$ and $\bhTheta_{\textsf{1-bit},\alpha}$ represents the probability estimators used in \eqref{prob:minA}, as described in Section \ref{sec:sim:miss}, and $t=0.05$ or $0.1$ denote the winsorized proportion for which $\beta$ is chosen.}
	\label{tab:Asimresr11n1}\par
	\centering
	\begin{threeparttable}
		\begin{tabular}{rlll}
			\hline
			($n_1,n_2$) = (600, 600) & $\text{RMSE}(\bhA_{\beta},\bAz)$ & Test Error & $r_{\bhA_{\beta}}$ \\
			\hline
			\textsf{Proposed\_$\bhTheta_{\textsf{Win},\beta}\_0.05$} &  1.5615 (0.0147) & 0.3005 (0.0062) & 65.28 (5.72) \\ 
			\textsf{Proposed\_$\bhTheta_{\beta}\_0.05$} & 1.5548 (0.0085) & 0.2996 (0.0034) & 54.98 (3.01) \\ 
			\textsf{Proposed\_$\bhTheta_{\textsf{Win},\beta}\_0.1$} & 1.5621 (0.0111) & 0.3013 (0.0046) & 63.68 (5.36) \\ 
			\textsf{Proposed\_$\bhTheta_{\beta}\_0.1$} & 1.5509 (0.0085) & 0.2983 (0.0034) & 53.13 (2.72) \\ 
			\textsf{Proposed\_$\bhTheta_{\alpha}$} & 1.5637 (0.0147) & 0.3010 (0.0061) & 65.63 (5.89) \\ 
			\textsf{Proposed\_$\bhTheta_{\textsf{1-bit},\textsf{Win},\beta}\_0.05$} & 1.5664 (0.0093) & 0.3028 (0.0037) & 62.76 (5.96) \\ 
			\textsf{Proposed\_$\bhTheta_{\textsf{1-bit},\beta}\_0.05$} & 1.5573 (0.0089) & 0.2996 (0.0036) & 61.80 (5.34) \\
			\textsf{Proposed\_$\bhTheta_{\textsf{1-bit},\textsf{Win},\beta}\_0.1$} & 1.5669 (0.0092) & 0.3032 (0.0037) & 62.78 (2.68) \\  
			\textsf{Proposed\_$\bhTheta_{\textsf{1-bit},\beta}\_0.1$} & 1.5540 (0.0089) & 0.2987 (0.0036) & 60.79 (3.01) \\ 
			\textsf{Proposed\_$\bhTheta_{\textsf{1-bit},\alpha}$} & 1.5612 (0.0097) & 0.3005 (0.0040) & 62.12 (4.76) \\ 
			\textsf{NW} & 1.9896 (0.2814) & 0.4676 (0.1341) & 167.67 (54.78) \\ 
			\textsf{KLT} & 2.2867 (0.0073) & 0.5951 (0.0026) & 1.00 (0.00) \\ 
			\textsf{MHT} & 1.6543 (0.0097) & 0.3432 (0.0041) & 51.20 (2.61) \\ 
			\hline
			($n_1,n_2$) = (800, 800) & $\text{RMSE}(\bhA_{\beta},\bAz)$ & Test Error & $r_{\bhA_{\beta}}$ \\
			\hline
			\textsf{Proposed\_$\bhTheta_{\textsf{Win},\beta}\_0.05$} & 1.4754 (0.0107) & 0.2669 (0.0041) & 88.58 (10.81) \\ 
			\textsf{Proposed\_$\bhTheta_{\beta}\_0.05$} & 1.4797 (0.0080) & 0.2714 (0.0030) & 71.79 (4.12) \\ 
			\textsf{Proposed\_$\bhTheta_{\textsf{Win},\beta}\_0.1$} & 1.4724 (0.0108) & 0.2664 (0.0042) & 86.25 (10.34) \\ 
			\textsf{Proposed\_$\bhTheta_{\beta}\_0.1$} & 1.4763 (0.0082) & 0.2704 (0.0031) & 67.08 (4.22) \\
			\textsf{Proposed\_$\bhTheta_{\alpha}$} & 1.4783 (0.0115) & 0.2676 (0.0041) & 88.92 (11.70) \\ 
			\textsf{Proposed\_$\bhTheta_{\textsf{1-bit},\textsf{Win},\beta}\_0.05$} & 1.4917 (0.0078) & 0.2743 (0.0030) & 83.51 (1.45) \\ 
			\textsf{Proposed\_$\bhTheta_{\textsf{1-bit},\beta}\_0.05$} & 1.4804 (0.0080) & 0.2705 (0.0031) & 82.60 (3.47) \\
			\textsf{Proposed\_$\bhTheta_{\textsf{1-bit},\textsf{Win},\beta}\_0.1$} & 1.4972 (0.0080) & 0.2765 (0.0031) & 81.64 (7.23) \\
			\textsf{Proposed\_$\bhTheta_{\textsf{1-bit},\beta}\_0.1$} & 1.4800 (0.0078) & 0.2708 (0.0030) & 74.89 (3.54) \\  
			\textsf{Proposed\_$\bhTheta_{\textsf{1-bit},\alpha}$} & 1.4790 (0.0099) & 0.2685 (0.0039) & 88.57 (9.56) \\  
			\textsf{NW} & 1.9515 (0.3625) & 0.4585 (0.1593) & 215.61 (82.24) \\ 
			\textsf{KLT} & 2.3447 (0.0064) & 0.6081 (0.0020) & 1.00 (0.00) \\ 
			\textsf{MHT} & 1.6067 (0.0086) & 0.3245 (0.0036) & 63.68 (3.02) \\ 
			\hline
		\end{tabular}
		\begin{tablenotes}
			\item [1]\footnotesize With $r_{\bMz}=11$, $r_{\bAz}$ = $11$, ($n_1,n_2$) = (600, 600), (800, 800) and SNR = $1$. The three existing methods are proposed respectively in \protect\citet{Negahban-Wainwright12}(\textsf{NW}), \protect\cite{Koltchinskii-Lounici-Tsybakov11}(\textsf{KLT}) and \protect\cite{Mazumder-Hastie-Tibshirani10}(\textsf{MHT})
		\end{tablenotes}
	\end{threeparttable}
\end{table}
\linsp

\setcounter{equation}{0} 

\section{Real data application}\label{sec:real}
In this section we demonstrate the proposed methodology by analyzing the Yahoo!~Webscope dataset (ydata-ymusic-user-artist-ratings-v1\_0) available at 

{\noindent \it http://research.yahoo.com/Academic\_Relations}.
It contains (incomplete) ratings from 15,400 users on 1000 songs.
The dataset consists of two subsets, a training set and a test set.
The training set records approximately 300,000 ratings
given by the aforementioned 15,400 users. Each song has at least 10 ratings.
The test set was constructed by surveying
5,400 out of these 15,400 users,
each rates exactly 10 songs
that are not rated in the training set.
The missing rates are $0.9763$ overall, $0.3520$ to $0.9900$ across
users, and $0.6372$ to $0.9957$ across songs. The non-uniformity of the missingness is
shown in Figure \ref{appfig:webuser} of Section \ref{appsec:real} in the supplementary material. In this experiment, we applied those
methods as described in Section \ref{sec:sim} to the training set and
evaluated the test errors based on the corresponding test set.
As there is no prior knowledge about true parameters $\alpha_1$ and $\alpha_2$, we suggest to choose $\alpha_1$ and $\alpha_2$ large enough, say $\alpha_1=100$ and $\alpha_2=100$, to ensure that the range covers all the missing probabilities. It was noted that $\bhTheta_\alpha$ is not sensitive to larger $\alpha$.

Table \ref{tab:real_data_comp} reports the root mean squared prediction errors, where
$\text{RMSPE}=\norm{\bW^{test}\circ(\bhA_{\beta}-\bY)}_{F}/(\sum_{i=1}^{n_1}\sum_{j=1}^{n_2}w_{ij}^{test})^{1/2}$ and $\bW^{test}$ is the indicator matrix of test set with the $(i,j)$th entry being $w_{ij}^{test}$. 
Note that \textsf{Proposed\_$\bhTheta_{\beta}$\_0.05} performs the best among all ten versions of proposed methods.
Besides, \textsf{Proposed\_$\bhTheta_{\alpha}$} also has much smaller root mean squared prediction error than the other eight versions of proposed methods. This may indicate that only
slight constraint is required for the probabilities estimator for this dataset.
Note that we cannot guarantee the optimal convergence rate or even asymptotic convergence in certain setting of missingness for \textsf{Proposed\_$\bhTheta_{\alpha}$}, see Section \ref{sec:asympA0} for details.  

With the separation of $\mu$, \textsf{Proposed\_$\bhTheta_{\alpha}$} is better than \textsf{Proposed\_$\bhTheta_{\textsf{1-bit},\alpha}$}; analogously, \textsf{Proposed\_$\bhTheta_{\beta}$\_$t$} is better than \textsf{Proposed\_$\bhTheta_{\textsf{1-bit},\beta}$\_$t$} with different constraint level $t$, same to \textsf{Proposed\_$\bhTheta_{\textsf{Win},\beta}$\_$s$} and \textsf{Proposed\_$\bhTheta_{\textsf{1-bit},\textsf{Win},\beta}$\_$s$} with different winsorization level $s$. 

As compared with the existing methods \textsf{NW}, \textsf{KLT} and \textsf{MHT},
our proposed methods perform significantly better in terms of root mean squared prediction errors,
and achieve as much as 25\% improvement when compared with Mazumder, Hastie and Tibshirani's method (the best among the three existing methods).
This suggests that a more flexible modeling of missing structure improves
the prediction power.

\linsps
\begin{table}[t!] 
	\small
	\centering
	\caption{Root mean squared prediction errors based on Yahoo!~Webscope dataset for the ten versions of the proposed method and the three existing methods proposed respectively in \protect\citet{Negahban-Wainwright12}(\textsf{NW}), \protect\cite{Koltchinskii-Lounici-Tsybakov11}(\textsf{KLT}) and \protect\cite{Mazumder-Hastie-Tibshirani10}(\textsf{MHT}).}
	\label{tab:real_data_comp}\par
	\begin{tabular}{cccc}
		\cline{1-3} 
		\hline
		& \textsf{Proposed\_$\bhTheta_{\textsf{Win},\beta}$\_0.05} & \textsf{Proposed\_$\bhTheta_{\beta}$\_0.05} & \textsf{Proposed\_$\bhTheta_{\textsf{Win},\beta}$\_0.1}  \\ 
		$\text{RMSPE}$ & 1.0396 & 1.0381 & 1.0476  \\
		\hline
		& \textsf{Proposed\_$\bhTheta_{\beta}$\_0.1} & 	\textsf{Proposed\_$\bhTheta_{\alpha}$} & \textsf{Proposed\_$\bhTheta_{\textsf{1-bit},\textsf{Win},\beta}$\_0.05} \\
		$\text{RMSPE}$ &  1.0490 & 1.0383 & 1.0831 \\
		\hline
		& \textsf{Proposed\_$\bhTheta_{\textsf{1-bit},\beta}$\_0.05} & \textsf{Proposed\_$\bhTheta_{\textsf{1-bit},\textsf{Win},\beta}$\_0.1} & \textsf{Proposed\_$\bhTheta_{\textsf{1-bit},\beta}$\_0.1} \\ 
		$\text{RMSPE}$ & 1.1091 & 1.0760 & 1.0523 \\ 
		\hline
		&	\textsf{Proposed\_$\bhTheta_{\textsf{1-bit},\alpha}$} & \textsf{NW} & \textsf{KLT}\\
		$\text{RMSPE}$ & 1.1065 & 1.7068 & 3.6334\\ 
		\hline 
		& \textsf{MHT} & \\
		$\text{RMSPE}$ & 1.3821 \\ 
		\hline  
	\end{tabular}
\end{table}
\linsp

\setcounter{equation}{0} 

\section{Concluding Remarks}\label{sec:conclude}
When the matrix entries are heterogeneously observed
due to selection bias, this heterogeneity should be taken into account.
This paper focuses on the problem of matrix completion under low-rank missing structure.
In the recovery of probabilities of observation, we
adopt a generalized linear model with a low-rank linear predictor matrix.
To avoid unnecessary bias, we introduce a separation of the mean effect $\mu$.
As the extreme values of probabilities may lead to unstable estimation of target matrix,
we propose an inverse probability weighting based method
with constrained probability estimates
and demonstrate the improvements
in empirical perspectives.
Our theoretical result
shows that the estimator of the high dimensional
probability matrix
can be embedded into the inverse probability weighting framework
without compromising the rate of convergence of
the target matrix (for an appropriately tuned $\beta>0$),
and reveals
a possible regime change in the tuning of
the constraint parameter ($\beta>0$ vs.~$\beta=0$).
In addition, corresponding computational algorithms are developed,
and a related algorithmic convergence result is established.
Empirical studies show the attractive
performance of the proposed methods as compared with existing matrix completion
methods.

\bibliographystyle{rss}  
\bibliography{MC_low_rank}

\end{document}